\def\year{2018}\relax
\documentclass[letterpaper]{article} 
\usepackage{aaai18}  
\usepackage{times}  
\usepackage{helvet}  
\usepackage{courier}  
\usepackage{url}  
\usepackage{graphicx}  
\usepackage{subcaption}
\usepackage{amsmath,amssymb}
\usepackage{balance}
\usepackage{multirow}

\newcommand{\eg}{e.g.}
\newcommand{\ie}{i.e.}
\newcommand{\etc}{etc.}
\newcommand{\bl}{\ensuremath{\mathcal{L}}}
\newcommand{\vect}[1]{\ensuremath{\mathbf{#1}}}

\DeclareMathOperator{\Tr}{Tr}

\usepackage{tikz}
\usetikzlibrary{plotmarks}
\usetikzlibrary{patterns}
\usepackage{pgfplots}
\pgfplotsset{compat=1.6}

\begin{document}
%
\title{End-to-End Video Classification with Knowledge Graphs}
\author{Fang Yuan, Zhe Wang, Jie Lin, Luis Fernando D'Haro, Kim Jung Jae, Zeng Zeng, Vijay Chandrasekhar\\ 
yfang@i2r.a-star.edu.sg, mark.wangzhe@gmail.com, luisdhe@i2r.a-star.edu.sg, jjkim@i2r.a-star.edu.sg \\
zengz@i2r.a-star.edu.sg, vijay@i2r.a-star.edu.sg\\
Institute for Infocomm Research, Singapore\\
}
\maketitle

\begin{abstract}
Video understanding has attracted much research attention especially since the recent availability of large-scale video benchmarks.
In this paper, we address the problem of multi-label video classification. We first observe that there exists a significant \emph{knowledge gap} between how machines and humans learn.
That is, while current machine learning approaches including deep neural networks largely focus on the representations of the given data, humans often look beyond the data at hand
and leverage external knowledge to make better decisions.
Towards narrowing the gap, we propose to incorporate external knowledge graphs into video
  classification. In particular, we unify traditional ``knowledgeless'' machine learning models and knowledge graphs in a novel end-to-end
  framework. The framework is flexible to work with most existing video
  classification algorithms including state-of-the-art deep models.
  Finally, we conduct extensive experiments on the largest public video dataset
  YouTube-8M. The results are promising across the board, improving mean average precision by up to 2.9\%.
\end{abstract}

\section{Introduction}

Since the advent of neural networks and deep learning, major breakthroughs have
been made in many artificial intelligence tasks ranging from computer vision to natural language
processing.
However, a significant \emph{knowledge gap} still exist between machine and
human intelligence. In particular, humans often relate to and make
use of semantic knowledge outside of the task-specific data to make better decisions.
On the other hand, most machine learning algorithms including state-of-the-art deep
methods, only focus on the representation of the given data,
without leveraging any external knowledge that could benefit the given task.

\begin{figure*}
\begin{subfigure}[t]{0.35\textwidth}
\subcaption{Image from Microsoft COCO\footnotemark}
\centering
\includegraphics[height=3.2cm]{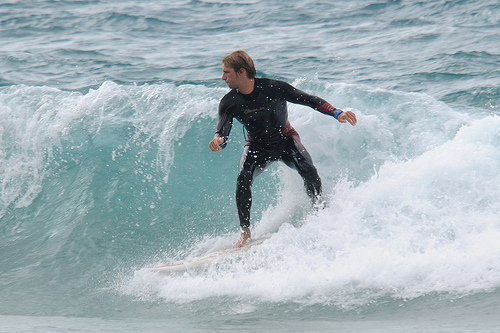}
\end{subfigure}%
\begin{subfigure}[t]{0.64\textwidth}
\subcaption{Video frames from YouTube\footnotemark}
\centering
\includegraphics[height=3.2cm]{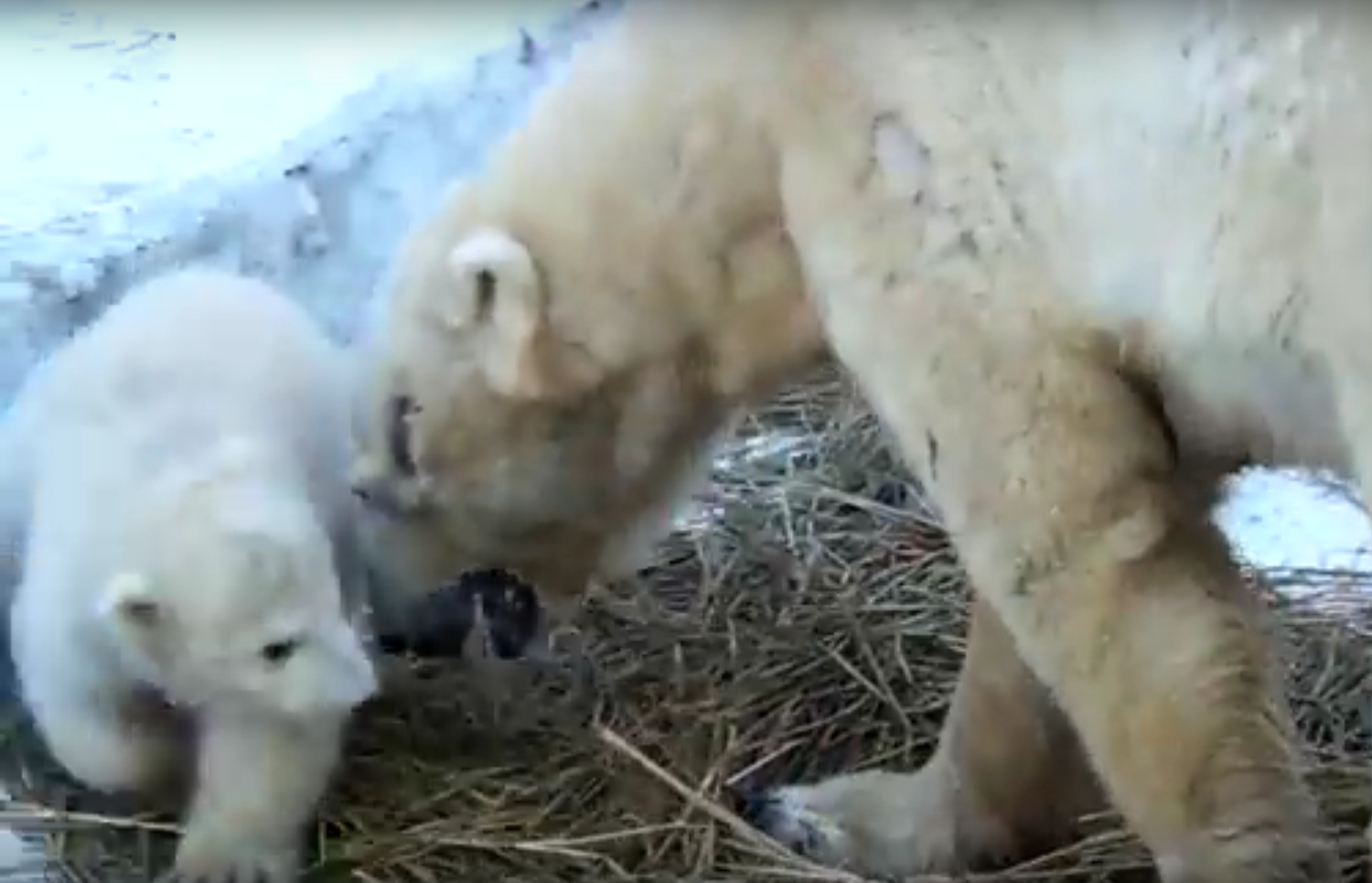}\hspace{2mm}
\includegraphics[height=3.2cm]{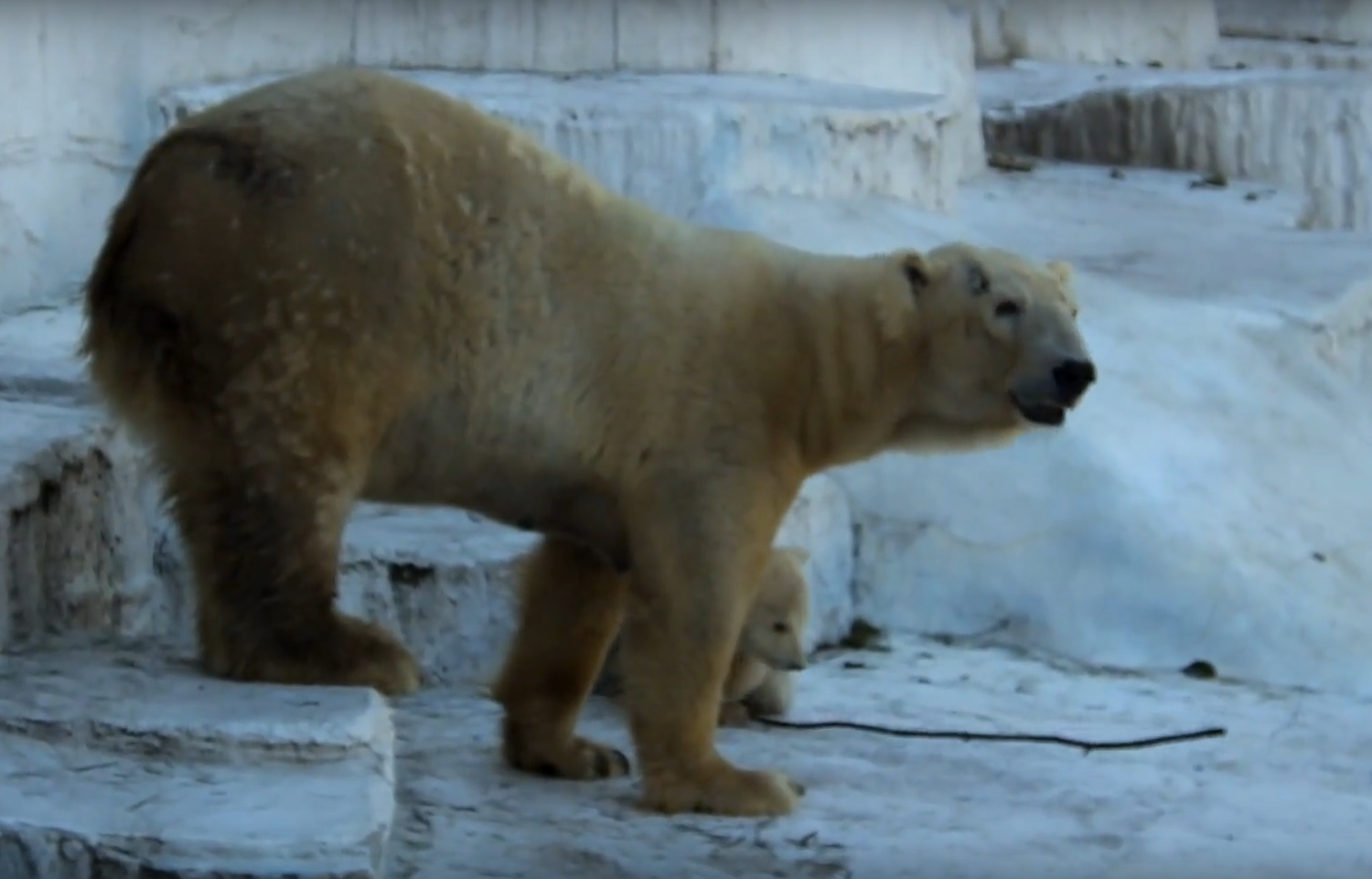}
\end{subfigure}
\caption{Bridging the knowledge gap between how humans and
machines learn in visual tasks: (a) Recognition of an obscured surfboard in an
image; (b)
Classification of zoo where the raw pixels in the video frames do not clearly
indicate a zoo.
\label{fig:surfing}}
\end{figure*}

Consider the surfing man example in
Figure~\ref{fig:surfing}(a). By only analyzing the pixels of the image (\ie,
given data), it is difficult to conclude that the man is on a surfboard since most of
it is obsecured by the waves. However, given the
knowledge that a man cannot stand freely on water and surfing is a typical
sport at sea, it is straightforward to identify the surfboard in
the picture. Note that the knowledge crucial to recognizing the
surfboard is external to the raw data. Likewise, in the zoo example in
Figure~\ref{fig:surfing}(b), with the knowledge that
man-made structures containing polar bears are most likely zoos, the video can
be correctly classified as zoo even though it is not evident from the appearance
of the structure in the frames.

In this paper, we study the problem of video classification.
In contrast to traditional ``knowledgeless'' models, we aim to design an
end-to-end ``knowledge-aware'' framework that can integrate external knowledge
into the learning process. The incorporation of knowledge is especially
critical to large-scale video classification benchmarks such as the recently
released YouTube-8M dataset \cite{abu2016youtube}, which presents two major
challenges \cite{wang2017truly}.
First, videos can be very diverse in nature, with vastly different topics
(\eg, sports, politics, entertainment, \etc) and genres (\eg, animation,
documentary, \etc). Second, the class distributions are highly imbalanced,
where majority of the classes have only very few instances.
Such diversity and imbalance makes the classes not easily separable based
only on features in the videos. Thus, external knowledge can play a
vital role in complementing the video features to attain higher classification
performance.

\begin{figure}
\centering
\includegraphics[scale=0.3]{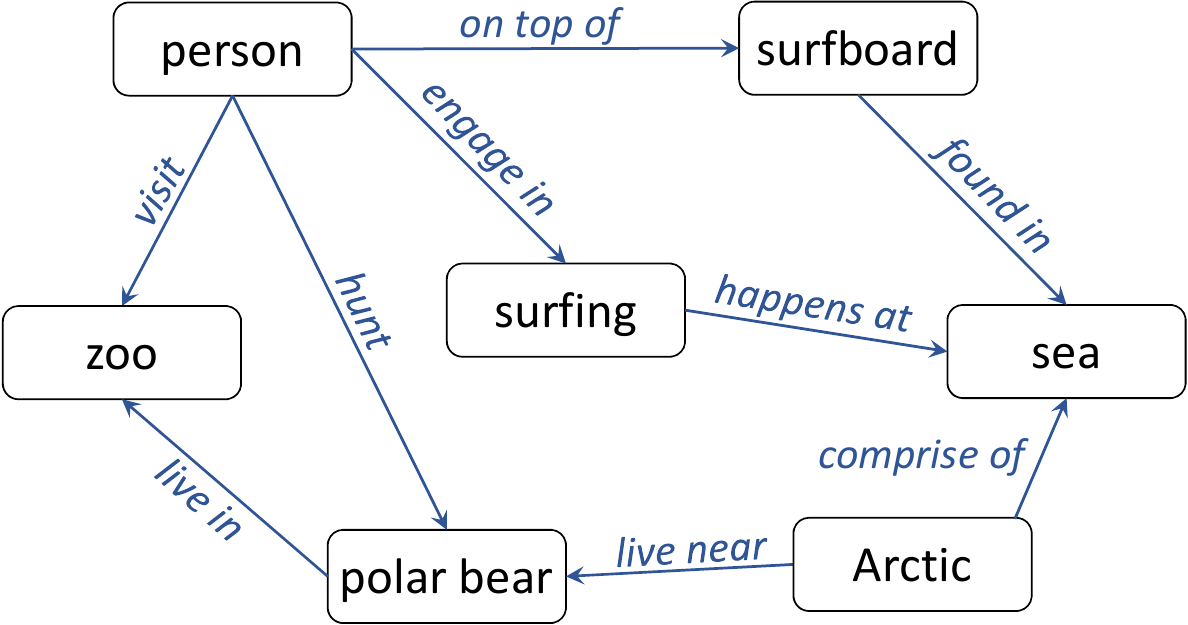}
\caption{Toy knowledge graph.\label{fig:toygraph}}
\end{figure}

More formally, knowledge is often represented as a \emph{knowledge
graph} \cite{DBLP:journals/semweb/Paulheim17}, modeling each real-world concept as a
node, and each semantic relationship between two concepts as an edge. A toy
knowledge graph is illustrated in Figure~\ref{fig:toygraph}. In
particular, the relationships ``person--on top of--surfboard'' and
``surfboard--found in--sea'' are likely to reinforce the recognition of
surfboard in Figure~\ref{fig:surfing}(a); similarly the relationship ``polar bear--live
in--zoo'' could help with the classification of zoo in
Figure~\ref{fig:surfing}(b).
While knowledge graphs have already seen widespread use in fields such as Web search and social
networks \cite{DBLP:conf/kdd/0001GHHLMSSZ14}, it has not been integrated
into visual tasks including video classification in a flexible and end-to-end fashion---Most existing knowledge-aware approaches are either
specific to a particular task and model, or applying external knowledge as a decoupled after-thought, which we will elaborate in related work.

Towards knowledge-aware video classification, we make the following
contributions in this paper.
\begin{itemize}
  \item We propose to incorporate external knowledge graphs into video
  classification, bridging the gap in existing state-of-the-art
  approaches.
  \item We unify knowledge graphs and machine learning including deep neural
  networks in a novel end-to-end learning framework, which is flexible to work with most existing learning models.
  \item We conduct extensive experiments on the largest public video dataset
  YouTube-8M, outperforming state-of-the-art methods by up to 2.9\% in mean average precision.
\end{itemize}

The remainder of this paper is organized as follows. First, we review related
work. Next, we present the proposed approach, followed by experimental evaluation. Finally, we conclude our
paper and lay out directions for future research.

\addtocounter{footnote}{-1}
\footnotetext{http://cocodataset.org/}
\stepcounter{footnote}\footnotetext{https://www.youtube.com/}

\begin{figure*}[tbp]
\centering
\includegraphics[width=0.99\textwidth]{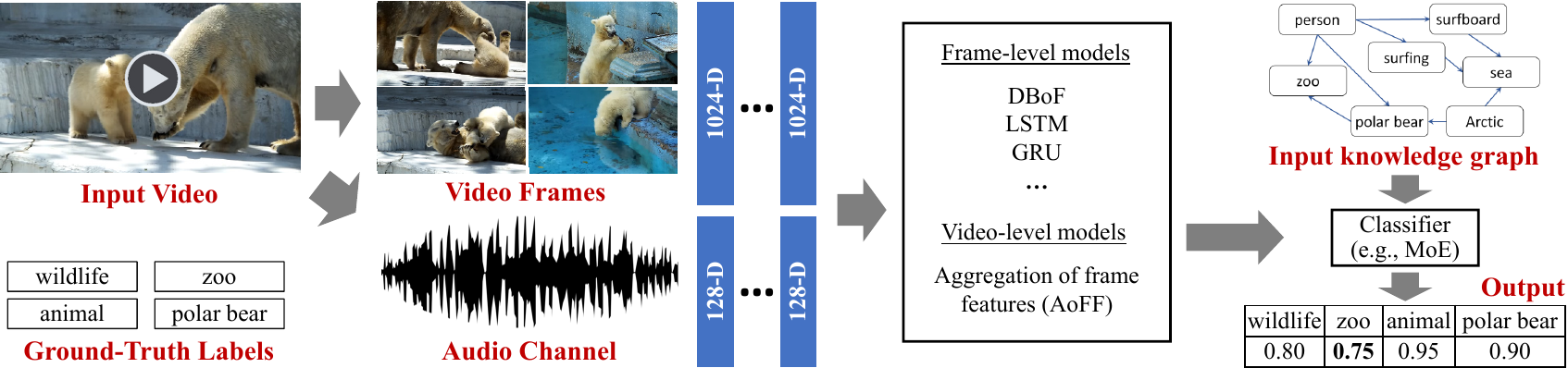}
\caption{Overall framework of the proposed end-to-end
learning with knowledge graphs (example video from YouTube).\label{fig:overall}}
\end{figure*}

\section{Related Work}
Video understanding has been an active research area in computer vision.
Significant progress has been made especially since the
release of large-scale benchmarks such as Sports-1M \cite{karpathy2014large},
YFCC-100M \cite{thomee2015new} and YouTube-8M \cite{abu2016youtube}.

The problem of video classification is usually addressed at frame or video
levels. The deep bag-of-frames (DBoF) \cite{abu2016youtube} is a typical
frame-level approach, inspired by various classic bag-of-words representations
\cite{laptev2008learning,wang2009evaluation}.
It feeds frame-level features from randomly sampled
input frames into a fully connected layer, whose parameters
are shared across the input frames.
Beyond a bag of frames, a video is naturally a temporal sequence of
frames, which can be modeled using a recurrent neural network.
Typically Long Short-Term Memory (LSTM) cells can be
employed to capture long-term dependencies in the temporal dimension
\cite{yue2015beyond}.
Furthermore, as an alternative to LSTM, Gated Recurrent Unit (GRU)
often achieves comparable if not better performance
\cite{chung2014empirical,chen2017aggregating}.
At video level, a fixed-length feature vector is often extracted from the
frames through simple aggregation (which we call ``aggregation of frame features'' or AoFF). As such, standard classifiers including
logistic regression and support vector machines can be adopted. In particular, the mixture of experts (MoE)
\cite{jordan1994hierarchical} classifier has shown superior empirical
performance on video-level representations \cite{abu2016youtube}.
In this model, a binary classifier is trained for each class, which is composed
of a set of ``experts'' or hidden states, and a softmax function is used to
model the probability of selecting an expert.
Apart from frame or video-level features directly extracted from the videos,
there is also initial success in exploiting text features associated with the videos,
such as the accompanied title and keywords in YouTube \cite{wang2017truly}.

All of the above methods are knowledgeless in the sense that they do not exploit external knowledge.
The use of external knowledge is emerging in some computer vision
tasks, including image classification
\cite{DBLP:conf/eccv/DengDJFMBLNA14}, motivation prediction \cite{DBLP:conf/cvpr/VondrickOPT16}, question answering
\cite{DBLP:conf/cvpr/WuWSDH16}, relationship extraction
\cite{DBLP:conf/eccv/LuKBL16}, as well as object detection
\cite{DBLP:conf/ijcai/FangKLTC17}. However, most of these works are task or
model specific, and thus cannot be easily applied to different scenarios.
While the recent work on object detection \cite{DBLP:conf/ijcai/FangKLTC17} can
work with any detection models, their proposed approach is not
end-to-end. Rather, it consists of two stages:
in the first stage, object localizations and class probabilities are obtained
using any existing model; in the second stage, the class probabilities are
re-optimized based on a knowledge graph. In particular, the use of knowledge in
the second stage is independent from the first stage, which means
there is a lack of feedback mechanism for the knowledge to directly improve
the parametrization of the existing model.

Finally, knowledge graph is a popular choice to represent external knowledge,
for capturing both concepts and their pairwise relationships. The
use of knowledge graphs have already demonstrated various degrees of success
in machine learning applications including  Web search and social media
\cite{DBLP:conf/kdd/0001GHHLMSSZ14}.
Quite a number of large-scale knowledge graphs are available commercially or in
open source, which are
generally constructed based on human curation
\cite{DBLP:journals/cacm/Lenat95}, crowdsourcing
\cite{liu2004conceptnet,DBLP:journals/ijcv/KrishnaZGJHKCKL17}, and distillation
from semi-structured
\cite{DBLP:conf/www/SuchanekKW07,DBLP:conf/semweb/AuerBKLCI07} or unstructured data \cite{DBLP:conf/aaai/CarlsonBKSHM10,DBLP:conf/wsdm/FangC11}.
The details of knowledge graph construction is beyond the scope of this work.

\section{Proposed Approach}

We describe our end-to-end knowledge-aware learning in this section,
starting with some preliminaries, followed
by our choice of knowledge representation, as well as the eventual
knowledge-aware classification.

\subsection{Preliminaries and notations}

Consider a set of pre-defined class labels $\bl=\{1,2,\ldots,L\}$ and a set of
videos $d$.
We address the multi-label classification problem for videos, where each video has one or
more ground-truth labels which form a subset of \bl. We assume a supervised
setting where some training videos with known ground-truth labels are available.
Given a test video with hidden ground truth, the task is to estimate a series
of probabilities $(p_1,p_2,\ldots,p_L)$ where $p_i$
represents the probability of label $i$ on the video. We can subsequently rank
the labels in descending probability and take the top few as the final output.

In this work, we further assume a knowledge graph. Many off-the-shelf knowledge graphs \cite{DBLP:journals/semweb/Paulheim17} exist for our purpose.
A knowledge graph is formally a graph $G=(V,E)$: $V$ is a set of vertices and $E$ is a set of edges between the vertices.
In the context of knowledge graph, each vertice represent a concept or class
label\footnote{We use the terms \emph{concept} and \emph{label}
interchangeably hereafter.}, and each edge represent a relationship between two
concepts.
A typical large-scale knowledge graph often contains millions or billions of concepts, and hundreds or thousands of different relationship types.

\subsection{Overall end-to-end framework}

The overall framework of our proposed end-to-end learning with knowledge graphs
is presented in Figure~\ref{fig:overall}. Given an input video, we can first
extract video and audio feature vectors from each frame.
Note that in YouTube-8M, the pre-extracted video and audio features per frame
consist of 1024 and 128 dimensions, respectively, as exemplified in the
diagram. The frame-by-frame feature vectors are then feed into either frame or
video-level models, to produce ultimate input into the classifier. As our main
novelty, in addition to accounting for features from the video instance, our
classifier further integrates a knowledge graph to narrow the knowledge gap
between traditional machine learning and human intelligence. As such, in our
running example, while the man-made structure is not clearly a zoo from the
frame pixels, we are still able to predict it with the help of a knowledge
graph, which reveals the strong semantic tie between polar bears and
zoos.

The proposed framework embodies two advantages. First, it enables the
incorporation of most existing video classification algorithms, including both
deep and shallow models. Thus, our framework can be highly flexible,
without being approach or task-specific. Second, the
unification with knowledge graphs happens within an end-to-end framework, which means external knowledge can directly influence the feature-based models in a feedback loop through mechanisms such as
backpropagation. In contrast, one recent approach for the related task of object
recognition \cite{DBLP:conf/ijcai/FangKLTC17} also draws input from
knowledge graphs. However, it is not end-to-end; it consists of two decoupled
stages where external knowledge is independent of the feature-based model. Due
to the lack of a feedback loop, their performance turns out to be unsatisfactory in video classification.

\subsection{Knowledge representation}

While external knowledge is commonly represented as graphs, knowledge graphs are
inherently  still symbolic and relational. Thus, quantifiable semantics must be
further extracted to enable integration with machine learning models which
typically operate over numerical representations.
The notion of \emph{semantic consistency} has been used
\cite{DBLP:conf/ijcai/FangKLTC17} to quantify the strength of semantic ties
between class labels. Generally two labels with high semantic
consistency suggests that they are likely to show up in
the same video. For instance, polar bear and zoo are two semantically consistent
concepts, whereas polar bear and volcano have weak or no semantic consistency.

We can encode semantic consistency in an
$L \times L$ matrix $S$, such that $S_{ij}$ represents the semantic
consistency between labels $i$ and $j$, $\forall ij \in \bl^2$. In particular,
$S_ij$ can be established based on the edges connecting the nodes representing
labels $i$ and $j$ on the knowledge graph.
Note that two nodes can be either directly connected by an edge (\eg, polar
bear--zoo), or indirectly through a path of edges (\eg, person--surfing--sea),
improving the generalization ability for concepts without any direct edge.
There can also exists multiple paths between two labels for
robustness.
Intuitively, between two nodes on the knowledge graph, when there are more paths
and these paths are shorter, their semantic consistency is stronger.

Random walk with restart
\cite{DBLP:conf/icdm/TongFP06} is a well-known method to realize the above
intuition. Starting from one node representing label $i$, we compute the
probability $R_{ij}$ of reaching another node representing label $j$ through
random walk.
The higher probability $R_{ij}$ implies that there are more and shorter paths
from $i$ to $j$ and thus the semantic consistency $S_{ij}$ is also higher. As
$R_{ij}\ne R_{ji}$ in general, but the semantic consistency matrix $S$ should
be symmetric in our context, we adopt the below definition follow the earlier
work \cite{DBLP:conf/ijcai/FangKLTC17}. We refer readers to existing work
\cite{DBLP:conf/icdm/TongFP06,DBLP:conf/icde/FangCL13,DBLP:journals/pvldb/ZhuFCY13}
on the computation of random walk probabilities $R_{ij}$.
\begin{align}
S_{ij} = S_{ji}=\sqrt{R_{ij}R_{ji}}
\end{align}

It is worth noting that semantic consistency can also be defined based on
the similarity of node embeddings, as enabled by recent representation learning
approaches on graphs \cite{grover2016node2vec,NIPS2013_5071}. However, our
proposed approach is orthogonal to the computation of semantic consistency,
which is beyond the scope of this paper.

Finally, for efficiency it is preferable to make the matrix $S$
sparser, by only focusing on the largest semantic consistency. To this end, we
consider the $K$-nearest neighbor (KNN) reduction for matrix $S$. A pair of
labels $i$ and $j$ are deemed KNN if $S_{ij}$ is one of the largest $K$ elements
in the $i$-th row or $j$-th row of $S$. Subsequently, we simply set
$S_{ij}=S_{ji}=0$ iff $i$ and $j$ are not KNN. The resulting matrix is much sparser, as it only encodes the strongest semantic consistency.

\subsection{Knowledge-aware classification}

Consider any classifier with a cost function $C$ and model parameters $\Theta$.
For a given video instance, we propose the following knowledge-aware cost function $K$,
where
$p_1,p_2,\ldots,p_L$ encode the label probabilities of the video and they are
functions of $\Theta$.
\begin{align}\label{eq:cost-pairwise}
K(\Theta)=C(\Theta)+\lambda \sqrt{\textstyle\sum_{i=1}^L
\sum_{j < i} S_{ij}(p_i-p_j)^2}
\end{align}

On the one hand, the original cost function $C$ captures the frame or video-level features,
whether they came from deep models or simple aggregation.
On the other hand, the new term here captures the semantics from the knowledge graph.
For a pair of labels $i$ and $j$, if $S_{ij}$ is large (\ie, the two labels have strong semantic consistency), minimizing
the cost function would force $p_i$ and $p_j$ to become similar. That is, it is likely that they either both appear in the video, or both not appear.
In contrast, if $S_{ij}$ is small (\ie, they are not semantically consistent), $p_i$ and $p_j$ become less constrained by the knowledge graph.
Note that the two cost terms, on the features and knowledge graph respectively, are balanced through a hyperparameter $\lambda \in (0,\infty)$.

\begin{figure}
\begin{subfigure}[t]{0.56\columnwidth}
\subcaption{Pairwise computations}
\centering
\includegraphics[scale=0.27]{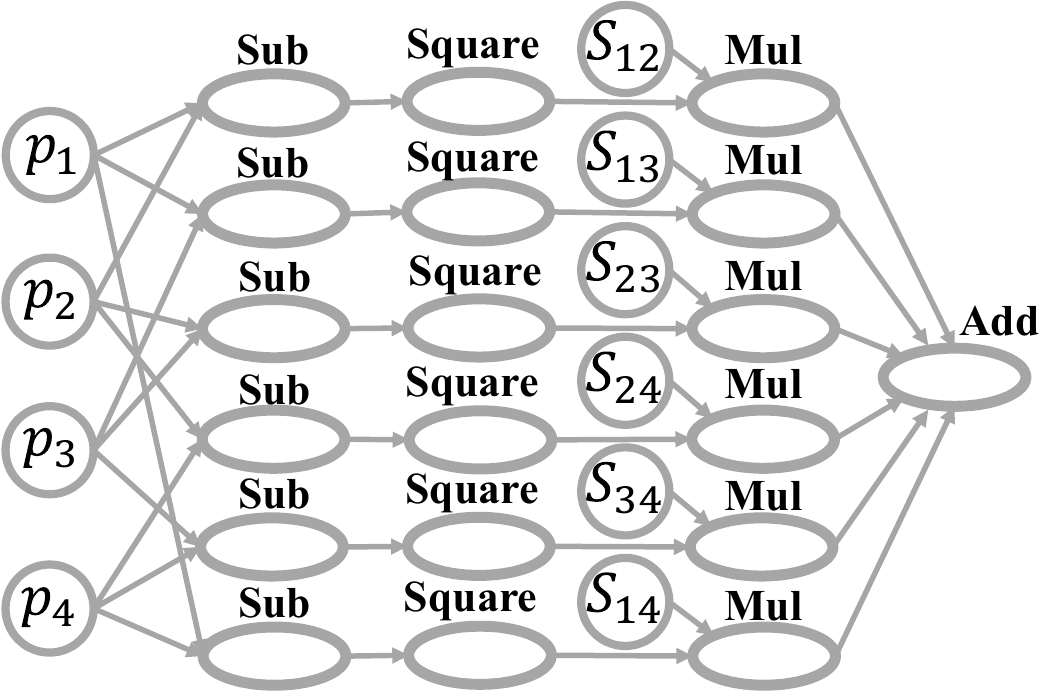}
\end{subfigure}%
\begin{subfigure}[t]{0.49\columnwidth}
\subcaption{Matrix computations}
\centering
\vspace{6mm}
\includegraphics[scale=0.27]{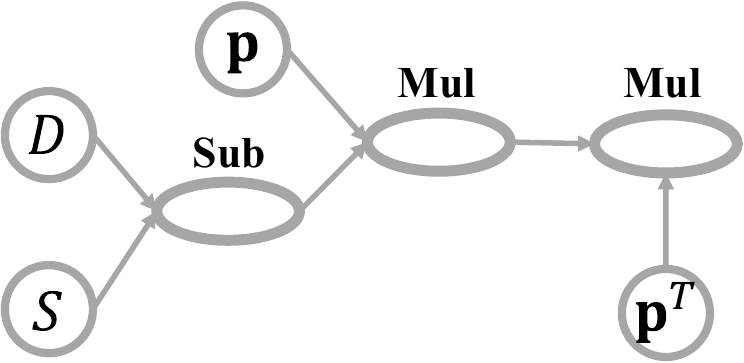}
\end{subfigure}
\caption{TensorFlow computation
graphs.\label{fig:tensorflow}}
\end{figure}

While the above formulation is intuitive, it is not practical for implementation
with standard libraries such as
TensorFlow\footnote{https://www.tensorflow.org/}.
In particular, TensorFlow operations are organized into a dataflow graph, as
illustrated in Figure~\ref{fig:tensorflow}(a) for the pairwise computation in
Equation~\eqref{eq:cost-pairwise} with $L=4$ and batch size $M=1$ (\ie, for a
single video).
Evidently, the dataflow graph would contain $O(L^2M)$ nodes, which is
not scalable in terms of the time required to construct this
graph, as well as the memory overhead incurred by storing the computation of all the intermediate nodes.

As such, we employ the Laplacian matrix transformation. It has been established
\cite{weiss2009spectral} that Equation~\eqref{eq:cost-pairwise} is equivalent to
the following:
\begin{align}
K(\Theta)=C(\Theta)+\lambda \sqrt{\Tr\left[\vect{p}(D-S)\vect{p}^T\right]}
\end{align}
where $D$ is a diagonal matrix such that $D_{ii} = \sum_{j=1}^L S_{ij}$ and
$\vect{p}=(p_1,p_2,\ldots,p_L)$ is a vector of label probabilities.
Note that $D-S$ is known as the Laplacian matrix. Using matrix computations, the dataflow graph is greatly
simplified as illustrated in Figure~\ref{fig:tensorflow}(b) with batch size
$M=1$. The total number of nodes simply become bounded by $O(M)$, improving the
scalability significantly.

Provided that the original cost
function $C$ and $\vect{p}$ are differentiable (which are generally true), our knowledge-aware cost
function is also differentiable, as follows. Thus, it can be optimized with the
gradient descent algorithm.
\begin{align}
\frac{\partial K(\Theta)}{\partial \Theta} & = \frac{\partial
C(\Theta)}{\partial \Theta} +
\lambda \frac{\partial \sqrt{\Tr\left[\vect{p}(D-S)\vect{p}^T\right]}}{\partial \vect{p}}\frac{\partial \vect{p}}{\partial \Theta} \nonumber\\
& = \frac{\partial C(\Theta)}{\partial
\Theta} + \lambda \frac{\vect{p}(D-S)}{\sqrt{\Tr\left[\vect{p}(D-S)\vect{p}^T\right]}}\frac{\partial
\vect{p}}{\partial \Theta}
\end{align}

\section{Empirical Evaluation}

In this section, we conduct empirical evaluations on the largest public video
classification benchmark to date, namely YouTube-8M.
We compare the performance of our approach against state-of-the-art video
classification models, and further investigate the impact of parameters on the
performance, and finally present some case studies to illustrate the reasons
that knowledge graphs can improve video classification.

\subsection{Experimental setup}

\subsubsection{Data}
We use the YouTube-8M benchmark\footnote{https://www.kaggle.com/c/youtube8m/data}, the largest public dataset
for multi-label video classification.
It contains over 7 million video instances and a diverse range of 4,716 classes
(entities), with an average of 3.4 labels per video. Pre-extracted and compressed
features at frame and video-levels are available, where the video and
audio features have 1024 and 128 dimensions, respectively.

We employ the off-the-shelf knowledge graph ConceptNet
5\footnote{http://conceptnet.io/}.
Following previous work \cite{DBLP:conf/ijcai/FangKLTC17}, we
only adopt its English subgraph, and remove self-loops and the so-called
``negative'' relationships (\eg, NotDesires, NotCapableOf,
Antonym and DistinctFrom). After these filtering steps,
we obtain a knowledge graph with 1.3 million concepts and 2.8 million
relationships. To further compute semantic consistency, we set the
random walk restarting probability to 0.15 as well.

To map the concepts in ConceptNet to class labels, we simply apply exact
string matching. As a result, 1,867 labels that have a path to at least one
other label are found in ConceptNet. To demonstrate the
advantage of using knowledge graphs, we only consider these 1,867 class labels,
which cover about 97\% the videos. Furthermore, these labels
account for almost 80\% of all label frequency. We emphasize that
obtaining better concept-class mapping for more coverage is not the focus of
this paper, and the current mapping already include the majority of the video
instances and label occurrences.

We use the given training set for training, and the given validation set for
testing since the ground truth of the original test set is not known.

\subsubsection{Evaluation metric}
For each test video, a ranked list of class labels is
produced, and we consider up to top 20 predictions per video for the following evaluation metrics.
\begin{itemize}
  \item Mean average precision (MAP): the mean value of the areas under the precision-recall curve of each video.
  \item Hit ratio (HIT): the percentage of test videos with the top one prediction belonging to the ground truth.
  \item Global average precision (GAP): area under the precision-recall curve over a global list of predictions consisting of all the predictions of all videos. 	
\end{itemize}

\subsubsection{Knowledgeless models}
Our framework is flexible to integrate knowledge graphs with different
``knowledgeless'' models (\ie, models without using external knowledge), including frame-level deep models and video-level models.
Thus, we consider four different state-of-the-art baseline models, namely,
AoFF, DBoF, LSTM and GRU. For the first three models \cite{abu2016youtube}, we
use the implementation by Google\footnote{https://github.com/google/youtube-8m};
for GRU \cite{miech2017learnable}, we use the implementation by Miech et al.\footnote{https://github.com/antoine77340/Youtube-8M-WILLOW}
More details of these models have been discussed in Related Work.
To train the models, we adhere to the setup in the two studies, as follows.
\begin{itemize}
  \item \textbf{AoFF}: learning rate 0.01, video-level model.
  \item \textbf{DBoF}: learning rate 0.01, 30 frames per video.
  \item \textbf{LSTM}: learning rate 1e-4, cell size 1024, all
  frames.
  \item \textbf{GRU}: learning rate 2e-4, cell size 1200, all
  frames.
\end{itemize}
Note that MoE classifier is used in all models, with 2 experts and 5 epochs.
We further set a batch size of 1024 for AoFF and 128 for the other three models.

\subsubsection{Knowledge-aware models}
We name our proposed end-to-end approach \textbf{E2E}. Each of the knowledgeless models (AoFF, DBoF, LSTM and GRU) can be coupled with knowledge graphs in our E2E framework.
We use $K=5$ for the KNN reduction of the semantic consistency matrix,
and $\lambda=0.01$ for the trade-off between feature-based cost and knowledge-based cost, which are generally robust values with stable performance.
We will vary these hyperparameters to study their impact on the performance as well.

We also compare to a previous knowledge-aware method
\cite{DBLP:conf/ijcai/FangKLTC17}. This method is originally designed for object detection in images, which can be adapted for multi-label video classification as well. It involves two stages,
where the first stage uses an existing knowledgeless model, and the second
stage uses a knowledge graph two re-optimize the output from the first stage.
Thus, the two stages are independent and their approach is not end-to-end.
We name this method \textbf{2STG}. We use $K=5$ for KNN as well, and choose $\epsilon=0.9$ which is found to be the best setting.

\subsection{Comparison of performance}

\begin{table*}[tb]
\center
\setlength\tabcolsep{4.5pt}
\renewcommand{\arraystretch}{1.3}
\begin{tabular}{c|ccc|ccc|ccc|ccc}
\hline
& \multicolumn{3}{c|}{\bf AoFF} & \multicolumn{3}{c|}{\bf DBoF} &  \multicolumn{3}{c|}{\bf LSTM} & \multicolumn{3}{c}{\bf GRU}\\
& MAP & HIT & GAP & MAP & HIT & GAP &  MAP & HIT & GAP &  MAP & HIT & GAP\\\hline
-			& 0.370 & 0.846 & 0.810 & 0.287 & 0.834 & 0.791 & 0.279 & 0.838 & 0.800 & 0.337 & 0.856 & 0.823\\\hline
\bf 2STG		& 0.364 & 0.841 & 0.804 & 0.286	& 0.830	& 0.787  & 0.275	 & 0.838	& 0.797	&0.331&	0.855&	0.819\\\hline
\bf \multirow{2}{*}{E2E}		& \bf 0.384			& \bf 0.849			& \bf 0.817			& \bf 0.301 		& \bf 0.847 		& \bf 0.794 		& \bf 0.296 		& \bf 0.854 		& \bf 0.808			& \bf 0.340 & \bf 0.857 & \bf 0.824\\[-0.5mm]
\bf 							& \small (+1.4\%)	& \small (+0.3\%)	& \small (+0.7\%)	& \small (+1.4\%)	& \small (+1.3\%)	& \small (+0.3\%)	& \small (+1.7\%)	& \small (+1.6\%)	& \small (+0.8\%)	& \small (+0.3\%)	& \small (+0.1\%)	& \small (+0.1\%)\\\hline
\end{tabular}
\caption{Performance comparison between E2E and 2STG across four state-of-the-art knowledgeless methods. The first row records the performance of the knowledgeless models;
the second row records the performance of 2STG that adopts the corresponding knowledgeless model in its first stage; the third row records the performance of E2E that couples with the corresponding knowledgeless model.
Bold entries represent the best value in each column.  \label{table:perf}}
\end{table*}

\begin{table*}[tb]
\center
\setlength\tabcolsep{4.3pt}
\renewcommand{\arraystretch}{1.3}
\begin{tabular}{c|ccc|ccc|ccc|ccc}
\hline
& \multicolumn{3}{c|}{\bf AoFF} & \multicolumn{3}{c|}{\bf DBoF} &  \multicolumn{3}{c|}{\bf LSTM} & \multicolumn{3}{c}{\bf GRU}\\
& MAP & HIT & GAP & MAP & HIT & GAP &  MAP & HIT & GAP &  MAP & HIT & GAP\\\hline
-			& 0.292	& 0.828 & 0.788 & 0.206	& 0.788	& 0.726 & 0.211	& 0.807	& 0.757 & 0.253	& 0.819	& 0.759\\\hline
\bf \multirow{2}{*}{E2E}		& 0.321			&  0.829			&  0.785			&  0.212 		&  0.801 		&  0.737 		&  0.232 		&  0.814 		&  0.767			&  0.259 &  0.826 &  0.776\\[-0.5mm]
\bf 							& \small\bf (+2.9\%)	& \small (+0.1\%)	& \small (-0.3\%)	& \small (+0.6\%)	& \small\bf (+1.3\%)	& \small\bf (+1.1\%)	& \small\bf (+2.1\%)	& \small (+0.7\%)	& \small\bf (+1.0\%)	& \small\bf (+0.6\%)	& \small\bf (+0.7\%)	& \small\bf (+0.7\%)\\\hline
\end{tabular}
\caption{Performance advantage of E2E across four state-of-the-art knowledgeless methods using only 10\% training data.
The percentage improvements of E2E are bolded if they are greater than or equal to the corresponding values in Table~\ref{table:perf}.
\label{table:perf10pct}}
\end{table*}

We first report the performance comparisons between the four knowledgeless
models and their respective knowledge-aware counterparts. Specifically, for
each knowledgeless model, we compare their results with those of both 2STG (previous
work) and E2E (our approach). The results are summarized in
Table~\ref{table:perf}. Note that we are only interested in comparing the values in each column, instead of comparing across different knowledgeless models.
Our approach E2E can achieve better performance every time, beating
the respective knowledgeless model by up to 1.7\% in MAP, 1.6\% in HIT and 0.8\%
in GAP. In contrast, 2STG performs poorly as it is not an end-to-end model. While 2STG can outperform knowledgeless models
for object detection on the Microsoft COCO \cite{DBLP:conf/eccv/LinMBHPRDZ14} and PASCAL VOC \cite{DBLP:journals/ijcv/EveringhamGWWZ10} datasets,
the two datasets involve only a restricted set of 80 and 20 classes, respectively. In contrast, our experiments deal with 1,867 classes over a much more complex and diverse range of topics, which could be the potential reason that the
decoupled two-stage method 2STG is unable to cope.

We further hypothesize that our knowledge-aware model has more advantage when the training set is smaller.
Intuitively, when there are fewer videos to learn from, the availability of external knowledge becomes ever more critical.
Using only 10\% training data, the performance of E2E with the four knowledgeless models is reported in Table~\ref{table:perf10pct}.
Not surprisingly, we observe slightly larger improvements than those using all training data, especially for the MAP and GAP metrics.

\subsection{Impact of parameters}

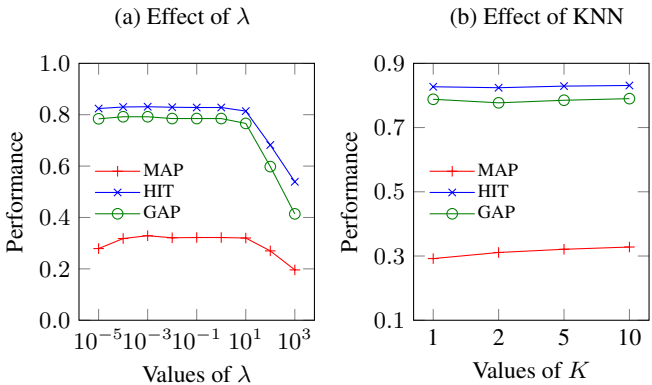
\begin{figure}[tbp]
\center
\begin{subfigure}[t]{0.56\columnwidth}
\center
\subcaption{Effect of $\lambda$}
\center
\hspace{-7mm}
\begin{tikzpicture}[scale=1][font=\small]
\begin{axis}[
		minor x tick style = transparent,
		ylabel=Performance,
		xlabel=Values of $\lambda$,
		enlarge x limits=0.1,
		ymin=0.0,
		ymax=1.0,
		ytick={0,0.2,0.4,0.6,0.8,1.0},
		xmode=log,
		xmin=1e-5,
		xmax=1000,
		xtick={1e-5,1e-3,0.1,10,1000},
		y tick label style={/pgf/number format/.cd,fixed,fixed zerofill,precision=1,/tikz/.cd},
		width=1\textwidth,
		height=5cm,
		legend style={at={(0.03,0.35)},anchor=south west,nodes={inner sep=1.0pt},fill=none,draw=none,font=\scriptsize},
		legend cell align=left,
		legend columns=1,
		xtick align=inside,
		]
\addplot[red,mark=+] coordinates {
(1e-5,	0.279)
(1e-4,	0.318)
(1e-3,	0.329)
(0.01,	0.321)
(0.1,	0.322)
(1,		0.322)
(10,	0.320)
(100,	0.270)
(1000,	0.196)
};
\addplot[blue,mark=x] coordinates {
(1e-5,	0.824)
(1e-4,	0.830)
(1e-3,	0.831)
(0.01,	0.829)
(0.1,	0.828)
(1,		0.828)
(10,	0.814)
(100,	0.682)
(1000,	0.539)
};
\addplot[green!50!black,mark=o] coordinates {
(1e-5,	0.784)
(1e-4,	0.792)
(1e-3,	0.792)
(0.01,	0.785)
(0.1,	0.785)
(1,		0.785)
(10,	0.766)
(100,	0.598)
(1000,	0.414)
};
\legend{MAP,HIT,GAP}
\end{axis}
\end{tikzpicture}
\end{subfigure}
\begin{subfigure}[t]{0.56\columnwidth}
\center
\subcaption{Effect of KNN}
\center
\hspace{-13mm}
\begin{tikzpicture}[scale=1][font=\small]
\begin{axis}[
		ylabel=Performance,
		xlabel=Values of $K$,
		ymin=0.1,
		ymax=0.9,
		ytick={0.1,0.3,0.5,0.7,0.9},
		symbolic x coords={1,2,5,10},
		y tick label style={/pgf/number format/.cd,fixed,fixed zerofill,precision=1,/tikz/.cd},
		width=1\textwidth,
		height=5cm,
		legend style={at={(0.03,0.35)},anchor=south west,nodes={inner sep=1.0pt},fill=none,draw=none,font=\scriptsize},
		legend cell align=left,
		legend columns=1,
		xtick align=inside,
		]
\addplot[red,mark=+] coordinates {
(1,		0.292)
(2,		0.311)
(5,		0.321)
(10,	0.328)
};
\addplot[blue,mark=x] coordinates {
(1,		0.827)
(2,		0.824)
(5,		0.829)
(10,	0.831)
};
\addplot[green!50!black,mark=o] coordinates {
(1,		0.788)
(2,		0.777)
(5,		0.785)
(10,	0.790)
};
\legend{MAP,HIT,GAP}
\end{axis}
\end{tikzpicture}
\end{subfigure}
\caption{Impact of parameters on the performance of E2E.\label{fig:paramimpact}}
\end{figure}

Next, we study the effect of parameters on the performance. There are two main
parameters for E2E: the trade-off $\lambda$ between the feature-based and
knowledge-based costs, and the choice of KNN for the semantic consistency
matrix. For brevity we only present their impact on AoFF,
as similar trends can be observed on other models.

In Figure~\ref{fig:paramimpact}(a), we vary $\lambda$ between 1e-5 and 1000,
while fixing KNN at $K=5$. Results show that the performance is generally stable
for a wide range of $\lambda$ between 1e-4 and 10. The performance only
deteriorates for very large values. Hence, it is robust to use $\lambda=0.01$ in
our experiments.

In Figure~\ref{fig:paramimpact}(b), we vary $K \in \{1,2,5,10\}$ for the choice
of KNN, while fixing $\lambda=0.01$. When we use larger $K$, there is a slight
increase in performance, especially in MAP, although the matrix $S$ becomes
denser and results in lower efficiency. Generally, using $K=5$ can
achieve a good balance between accuracy and efficiency.

\subsection{Result analysis}

\begin{table*}[tb]
\center
\renewcommand{\arraystretch}{1.3}
\resizebox{0.99\textwidth}{!}{%
\begin{tabular}{r|l|l|l|l}
\hline
& \bf Ground truth & \bf AoFF rank & \bf E2E rank & \bf Related concepts in the same video\\\hline
1 & fashion & 20 & 1 ($\uparrow$19) & hairstyle, bollywood, cosmetics \\
2 & origami & 20+ & 1 ($\uparrow$19+) & paper, toy\\
3 & amusement park & 20+ & 1 ($\uparrow$19+) & food, roller coaster, train\\
4 & disc jockey & 5 & 2 ($\uparrow$3) & nightclub, dance, album, guitar\\
5 & food, drink & 1, 5 & 1, 2 ($\uparrow$3) & recipe, cocktail, juice, cooking, bartender, bottle\\
6 & camera, photography & 4, 9 & 1 ($\uparrow$3), 4 ($\uparrow$5) & gadget, camera lens, smart phone\\
7 & hunting, deer & 4, 20+ & 1 ($\uparrow$3), 6 ($\uparrow$14+) & forest, tree, plant, animal, weapon\\
8 & vehicle, tool, drill & 2, 4, 20+ & 1 ($\uparrow$1), 2 ($\uparrow$2), 6 ($\uparrow$14+) & car, metalworking\\
9 & concert, lighting, festival & 2, 5, 16 & 1 ($\uparrow$1), 2 ($\uparrow$3), 8 ($\uparrow$8) & dance, album, Ibiza\\
10 & furniture, couch, bed, chair & 1, 3, 11, 20+ & 1, 2 ($\uparrow$1), 6 ($\uparrow$5), 4 ($\uparrow$16+) & living room, home improvement, house, television\\\hline
\end{tabular}}
\caption{Example videos that knowledge graphs can help with learning. Each row describes a video, where  ``AoFF rank'' and ``E2E rank'' columns indicate the rank position of the ground truth label in the output of AoFF and E2E, respectively;
20+ means the ground truth is not found in the top 20; $\uparrow$ indicates the number of positions moved up in E2E output as compared to AoFF; related concepts
are listed if they have high semantic consistency with the ground truth and they are within top 20 of both AoFF and E2E.\label{table:casepositive}}
\end{table*}

\begin{table*}[tb]
\center
\renewcommand{\arraystretch}{1.3}
\resizebox{0.99\textwidth}{!}{%
\begin{tabular}{r|l|l|l|l}
\hline
& \bf Ground truth & \bf AoFF top & \bf E2E top & \bf Other concepts in the same video\\\hline
1 & telescope & \bf telescope & \it vehicle & \textbf{camera}, \textit{car}, \textit{boat}, \textit{bicycle}, \textit{motorcycle}  \\
2 & transistor & \bf transistor & \it vehicle &  \textbf{antenna}, \textit{train}, \textit{car}  \\
3 & running, marathon & \textbf{running}, \textbf{hiking} & \textit{mountain}, \textit{nature} &  \textbf{climbing}, \textbf{walking}, \textit{mountain pass}, \textit{trail}, \textit{lake} \\
4 & gardening, plant & \textbf{plant}, \textbf{gardening} & \textit{food}, \textit{news program} & \textbf{tree}, \textit{agriculture}, \textit{cooking}, \textit{television} \\
5 & banknote, money, dollar & banknote, dollar, money & paper, \textit{animation}, \textit{guitar} & \textit{manga}, \textit{art}, \textit{festival}, \textit{musician}\\\hline
\end{tabular}}
\caption{Example videos where knowledge graphs can hurt performance. Each row describes a video, where ``AoFF top'' and ``E2E top'' columns indicate the top prediction(s) of AoFF and E2E, respectively;
other concepts are listed if they have high semantic consistency with the top prediction(s) of either AoFF or E2E, and they are within top 20 of both AoFF and E2E; bold entries are
a group of concepts with strong mutual semantic consistency; likewise for italic entries.
\label{table:casenegative}}
\end{table*}

Finally, we conduct a more in-depth analysis of the results, using AoFF as the knowledgeless model.
At an aggregate level, we observe better predictions for 24.9\% of the videos after incorporting external knowledge with E2E,
whereas we witness worse predictions for 8.9\% of the videos. The remaining videos have no change in their predictions.
Given that the number of videos with better results are almost three times of the videos with worse results, our approach E2E does bring in net benefits,
consistent with the quantitative evaluation reported earlier.

We further zoom into some specific examples to understand the reasons behind the improvement.
In Table~\ref{table:casepositive}, we illustrated 10 videos where knowledge graphs can help with learning.
In these cases, E2E significantly improves the rank positions of the ground truth labels over AoFF, due to the evidence of the related concepts in the same video.
For instance, in example \#2, the concept of origami is semantically consistent with paper and toy, which helps E2E to identify origami correctly.
On the contrary, even though AoFF also recognizes the concepts paper and toy in the same video, it has no idea that those concepts are related to origami, especially when there are not enough training videos
involving origami, paper and toy.

\vspace{1mm}

Finally, we investigate some negative cases in Table~\ref{table:casenegative}, where knowledge graphs can hurt the performance.
In these cases, E2E often makes overgeneralizations based on the related concepts. In these videos, there exist an overwhelm of concepts that are semantically consistent to the incorrect top predictions
by E2E, whereas the concepts that are consistent with the ground truth are much fewer (\eg, \#1 and \#4) or even non-existent (\eg, \#5).
The root cause is that E2E treats each related concepts uniformly. However, in an ideal solution, we should only focus on the concepts related to the central theme of the video.
We leave the study of such ``focus'' concepts as potential future work.

\section{Conclusion}

\balance

In this paper, we studied the multi-label video classification problem. In
particular, we observed the knowledge gap between machine and human
intelligence. Towards bridging this gap, we proposed to utilize external knowledge graphs
for video classification, unifying machine learning including deep neural
networks with knowledge graphs in a novel end-to-end framework. Extensive
experiments on the largest public benchmark YouTube-8M showed the superior performance of our approach, outperforming
state-of-the-art knowledgeless models by up to 2.9\% in MAP among other metrics. Finally, we analyzed some case studies to understand the scenarios in which knowledge graphs can or cannot help.

As future work, we plan to extract features from knowledge graphs and directly
incorporate them into the deep neural networks. Moreover, it is also worth investigating that how we can identify focus concepts
that are related to the central theme of a video.

\newpage

\balance
\bibliography{ref}
\bibliographystyle{aaai}

\end{document}